\documentclass[12pt]{article}

\usepackage{amsmath}
\usepackage[justification=centering,margin=10pt,font=small,labelfont=bf]{caption}
\usepackage{graphicx}

\usepackage{apacite}
\usepackage[comma,authoryear,round]{natbib}
\begin{document}

\noindent UNBIASED VARIABLE IMPORTANCE FOR RANDOM FORESTS
\vskip 3mm

\vskip 5mm
\noindent Markus Loecher

\noindent  Berlin School of Economics and Law, 

\noindent 10825 Berlin, Germany

\noindent markus.loecher@hwr-berlin.de

\vskip 3mm

\noindent ABSTRACT

The default variable-importance measure in random Forests, Gini importance, has been shown to suffer from the bias of the underlying Gini-gain splitting criterion.
While the alternative \textit{permutation importance} is generally accepted as a reliable measure of variable importance, it is also computationally demanding and suffers from other shortcomings.
We propose a simple solution to the misleading/untrustworthy Gini importance which can be viewed as an overfitting problem: we compute the  loss reduction on the out-of-bag instead of the in-bag training samples.



\section{Variable importance in trees}

Variable importance is not very well defined as a concept. 
Even for the case of a linear model with $n$ observations, $p$ variables and the standard $n >> p$ situation, there is no theoretically defined variable importance metric in the sense of a parametric quantity that a variable importance
estimator should try to estimate \citep{gromping2009variable}.
Variable importance measures for random forests have been receiving increased attention in bioinformatics, for instance to select a subset of genetic markers relevant for the prediction of a certain disease.
They also have been used as screening tools \citep{diaz2006gene,menze2009comparison}
in important applications highlighting the need for reliable and well-understood feature importance
measures.

The default choice in most software implementations \citep{randomForest2002,pedregosa2011scikit} of random forests \citep{Breiman2001} is the \textit{mean decrease in impurity (MDI)}. The MDI of a feature is computed as a (weighted) mean of the individual trees' improvement
in the splitting criterion produced by each variable. A substantial shortcoming of this default measure is its evaluation on the in-bag samples which can lead to severe overfitting \citep{kim2001classification}. It was also pointed out by \cite{Strobl2007a} 
that \textit{the variable importance measures of Breiman's original Random Forest method ... are not reliable in situations where potential predictor variables vary in their scale of measurement or their number of categories}.\\
There have been multiple attempts at correcting the well understood bias of the Gini impurity measure both as a split criterion as well as a contributor to importance scores, each one coming from a different perspective. \\
\cite{strobl2007unbiased} derive the exact distribution of the maximally selected Gini gain along with their resulting p-values by means of a combinatorial approach. 
\cite{shih2004variable} suggest a solution to the bias  for the case of regression trees as well as binary classification trees \citep{shih2004note} which is also based on p-values. Several authors \citep{loh1997split, hothorn2006unbiased} argue that the criterion for split variable and split point selection should be separated.

An idea that is gaining quite a bit of momentum is to add so-called pseudo variables to a dataset, which are permuted versions of the original variables and can be used to correct for bias \citep{sandri2008bias}. 
Recently,  a modified version of the Gini importance called Actual Impurity Reduction (AIR) was proposed \cite{nembrini2018revival} that is faster than the original method proposed by Sandri and Zuccolotto with almost no overhead over the creation of the original RFs and available in the R package \textit{ranger}  \citep{wright2015ranger,wright2017package}.
After submission of this article, the following two papers using OOB samples to compute a debiased version of the Gini importance  \citep{li2019debiased,zhou2019unbiased}  came to the authors' atttenion.
  
\noindent 
We use the well known titanic data set to illustrate the perils of putting too much faith into the Gini importance which is based entirely on training data - not on OOB samples - and makes no attempt to discount impurity decreases in deep trees that are pretty much frivolous and will not survive in a validation set.\\
In the following model we include \textit{passengerID} as a feature along with the more reasonable \textit{Age}, \textit{Sex} and \textit{Pclass}.
Figure~\ref{fig:titanic1} below show both measures of variable importance and (maybe?) surprisingly \textit{passengerID} turns out to be ranked number $3$ for the Gini importance (MDI). This troubling result is robust to random shuffling of the ID.

\begin{figure}[htbp]
  \centering
  \includegraphics{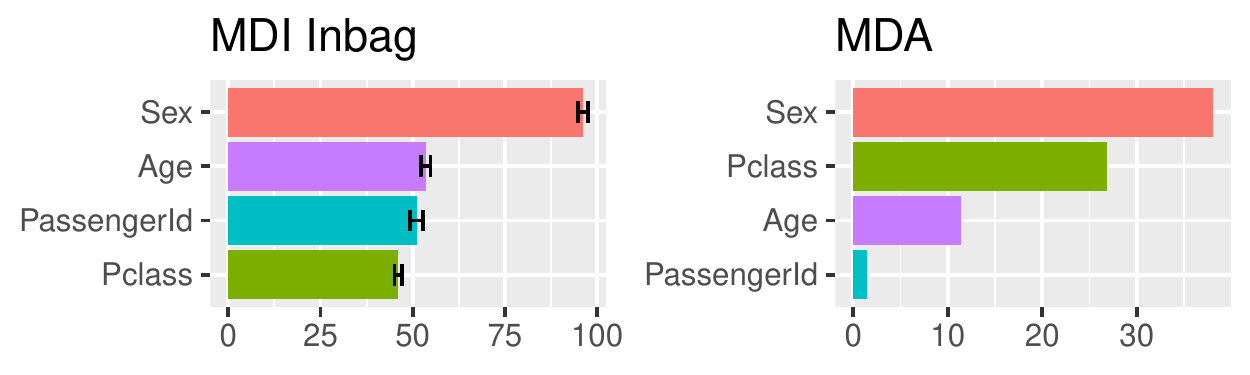}
  \caption{Mean decrease impurity  (MDI, left panel) versus permutation importance  (MDA, right panel)  for  the Titanic data.}
  \label{fig:titanic1}
\end{figure}
The permutation based importance (MDA, right panel) is not fooled by the irrelevant ID feature. This is maybe not unexpected as the IDs should bear no predictive power for the out-of-bag samples.

\noindent There appears to be broad consenus that random forests rarely suffer from “overfitting” which plagues many other models. (We define overfitting as choosing a model flexibility which is too high for the data generating process at hand resulting in non-optimal performance on an independent test set.) By averaging many (hundreds) of separately grown deep trees -each of which inevitably overfits the data - one often achieves a favorable balance in the bias variance tradeoff.
For similar reasons, the need for careful parameter tuning also seems less essential than in other models.
We point out that random forests’ immunity to overfitting is restricted to the predictions only and not to the default variable importance measure.

\section{OOB impurity reduction}

As is standard practice in statistical modeling and machine learning optimization, any attempt to generalize performance or other metrics needs to be based on validation data that were \textbf{not} used in the training stage.
We propose the mean decrease in loss evaluated on the OOB parts of the data  as a simple alternative to the computationally expensive permutation importance. This approach (i) is in close analogy to MDI (which is computed on the training data), (ii) should be easily implemented in most software libraries and (iii) leads to no bias toward variables with high cardinality.
As candidates for appropriate loss functions, we explored \textit{log loss}, \textit{misclassification rate} and  \textit{MDI}$_{OOB}$ but eventually adapted a \textit{penalized Gini impurity} which combines inbag and out-of-bag samples in a novel way.
It is easy to see that just redefining the information gain based on the Gini impurities of the OOB samples
\[
G_{OOB}(m) = 2 \hat{p}_{OOB}(m) \cdot \left(1- \hat{p}_{OOB}(m) \right)
\]
is not a good measure for the quality of a split as it ignores the correlation with the actual labels from the training data. 
It would be misleading to assign a low impurity to a node with e.g. $\hat{p}_{OOB}=0.9$ if the predictions of the training data were drastically different, e.g. $\hat{p}_{inbag}=0.2$. In such a case we would not want to boost the contribution of the variable which led to this split.

Our main idea is to increase the impurity $I(m)$ for node $m$ by a  penalty that is proportional to the difference $\Delta=(\hat{p}_{OOB} - \hat{p}_{inbag})^2$:
 \[
 PG_{OOB}^{\alpha,\lambda} = \alpha \cdot I_{OOB}   + (1-\alpha) \cdot  I_{inbag}  + \lambda \cdot (\hat{p}_{OOB} - \hat{p}_{inbag})^2
 \] 
The exact details of the actual loss functions depend on the following conceptual design choices:
\begin{itemize}
\item Symmetry: should one weigh train and test data equally or put more faith into the test impurity? 
\item Maximum Uncertainty: should one allow the discrepany $\Delta$ to increase the loss beyond its univariate maximum or keep the total loss bounded by it no matter how different $\hat{p}_{OOB}$ and $\hat{p}_{inbag}$.
\end{itemize}
The answers to both questions also affect the likelihood of negative importance scores which may or may not be desired, an issue that we will reflect upon later. 
The following four \textbf{penalized Gini impurities}\footnote{(0): $\alpha=1, \lambda=0$, (1): $\alpha=1, \lambda=1$, (2): $\alpha=0.5, \lambda=1$, (3): $\alpha=0.5, \lambda=0.5$} will be evaluated in this paper:
\begin{align}
PG_{oob}^{(0)} & = 2 \cdot  \hat{p}_{oob} \cdot (1- \hat{p}_{oob} )  \label{eq:PG0} \\
PG_{oob}^{(1)} & = 2 \cdot  \hat{p}_{oob} \cdot (1- \hat{p}_{oob} )   + (\hat{p}_{oob} - \hat{p}_{in})^2 \\
PG_{oob}^{(2)} & = \hat{p}_{oob} \cdot (1- \hat{p}_{oob} )  +   \hat{p}_{in} \cdot (1- \hat{p}_{in} )  + (\hat{p}_{oob} - \hat{p}_{in})^2 \\
PG_{oob}^{(3)} & = \hat{p}_{oob} \cdot (1- \hat{p}_{oob} )  +  \hat{p}_{in} \cdot (1- \hat{p}_{in} ) + \frac{1}{2} \cdot  (\hat{p}_{oob} - \hat{p}_{in})^2
\end{align}
The last two are clearly symmetric in $\hat{p}_{OOB}, \hat{p}_{inbag}$, whereas $PG_{OOB}^{(1)}$ does not take into account the inbag impurity of the node. Measures (1) and (2) ``over-penalize'', i.e. their maximum values are $1 > G_{OOB}^{max}$, while the last measure does not allow the $\Delta$ term to increase the impurity beyond $0.5 = G_{OOB}^{max}$.
We mention in passing that the terms \textit{inbag/OOB} are synonymous with \textit{train/test} subsets.

Our R package \textbf{rfVarImpOOB} \citep{rfVarImpOOB2019} computes $(1)-(3)$ for the current \textbf{randomForest} library \citep{randomForest2002}. The code is written as an external wrapper and hence slow. While it would be straightforward to parallelize the pass over the individual trees, our hope is that the authors of \citep{wright2015ranger,wright2017package,randomForest2002,pedregosa2011scikit} would adapt these importance scores into the $C$ code base.

\begin{figure}[htbp]
  \centering
  \includegraphics[width=\textwidth]{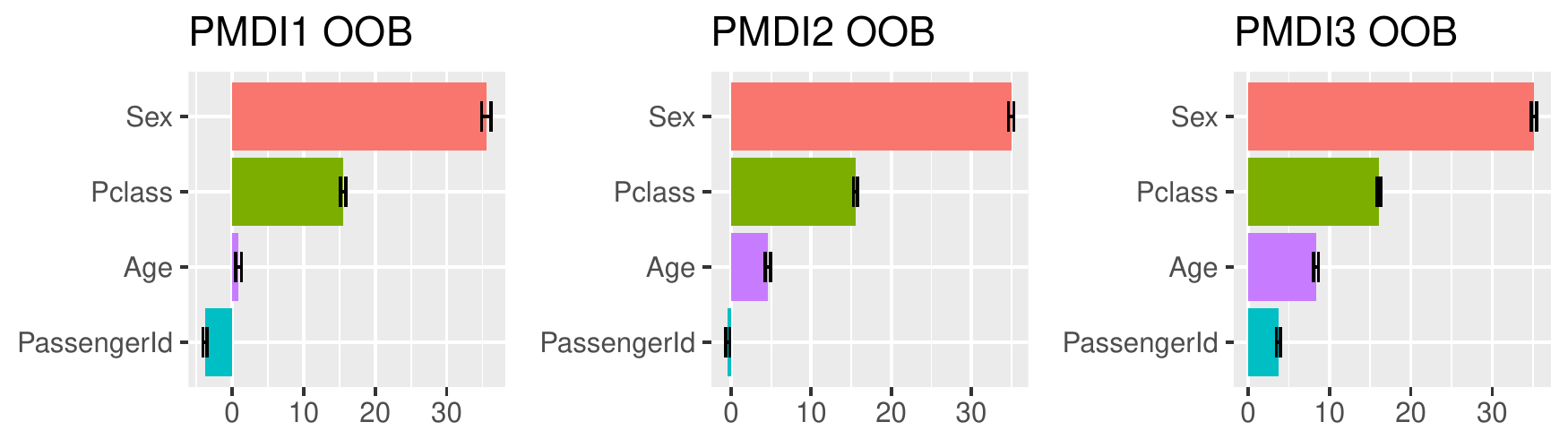}%
  \caption{Comparison of the three penalized (Gini) mean decrease impurities  (PMDI) eqns. $(1)-(3)$ for the Titanic data.  $PG_{OOB}^{(1)}$ penalizes overfitting individual trees the strongest, leading to negative scores for uninformative variables.  }
  \label{fig:titanic2}
\end{figure}

Figure~\ref{fig:titanic2} compares the three measures for the same random forest model fitted to the Titanic data.
While there is no objective ground truth to compare with, the left 2 panels assign effectively zero importance to \textit{PassengerId} compared to $PG_{OOB}^{(3)}$ which penalizes differences between train/test the least. The negative scores for \textit{PassengerId} from $PG_{OOB}^{(3)}$ can be interpreted as a measure of extreme overfitting in the tree building process. 
Note that the valid split contributions of \textit{Age} are overcompensated by the large number of splits that do not hold up on the validation set.
If negative importance scores are deemed difficult to communicate, it would be appropriate to truncate them at zero.
 
\subsection{Application to C-to-U conversion data}
 
\begin{figure}[htbp]
  \centering
  \includegraphics[width=\textwidth]{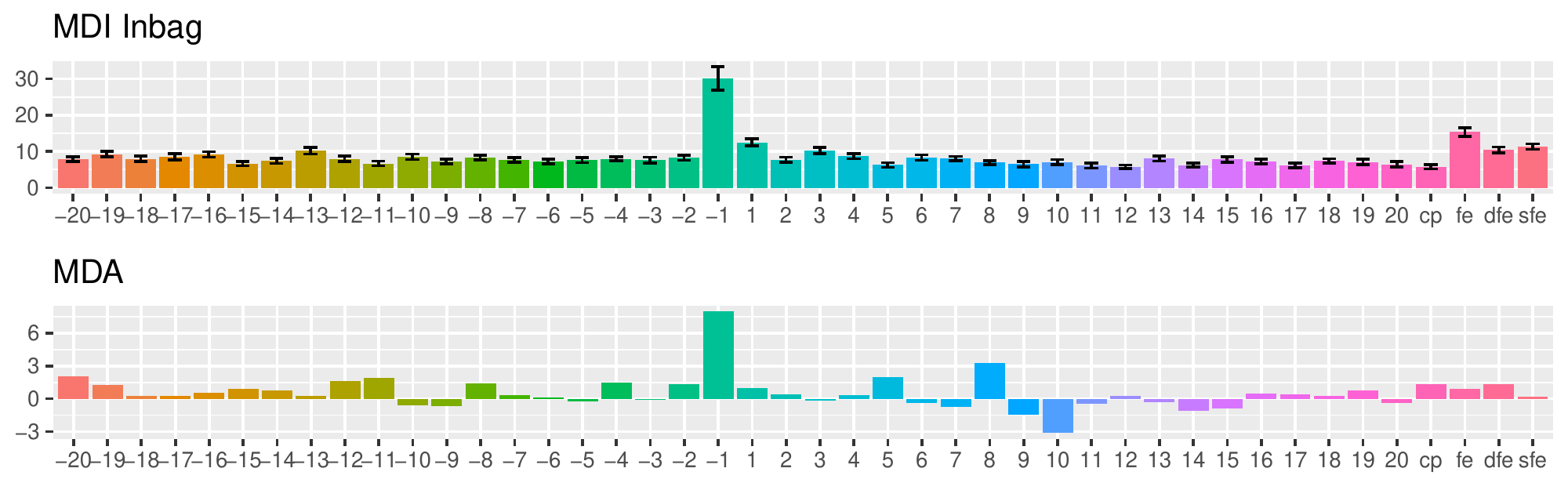}%
  \caption{Mean decrease impurity  (MDI, left panel) versus permutation importance  (MDA, right panel)  for  the Arabidopsis data. The random column \textit{sfe} achieves the $4$th highest rank for the MDI score. 
  Also note the many largely negative scores for the permutation accuracy which gives support for a similar behavior of $PG_{OOB}^{(1)}$. In each plot the positions -20 through 20 indicate the nucleotides flanking the site of interest, and the last three bars on the right refer to the codon position (cp), the estimated folding energy (fe) and the difference in estimated folding energy (dfe)}
  \label{fig:arabidopsis1}
\end{figure}

\begin{figure}[htbp]
  \centering
  \includegraphics[width=\textwidth]{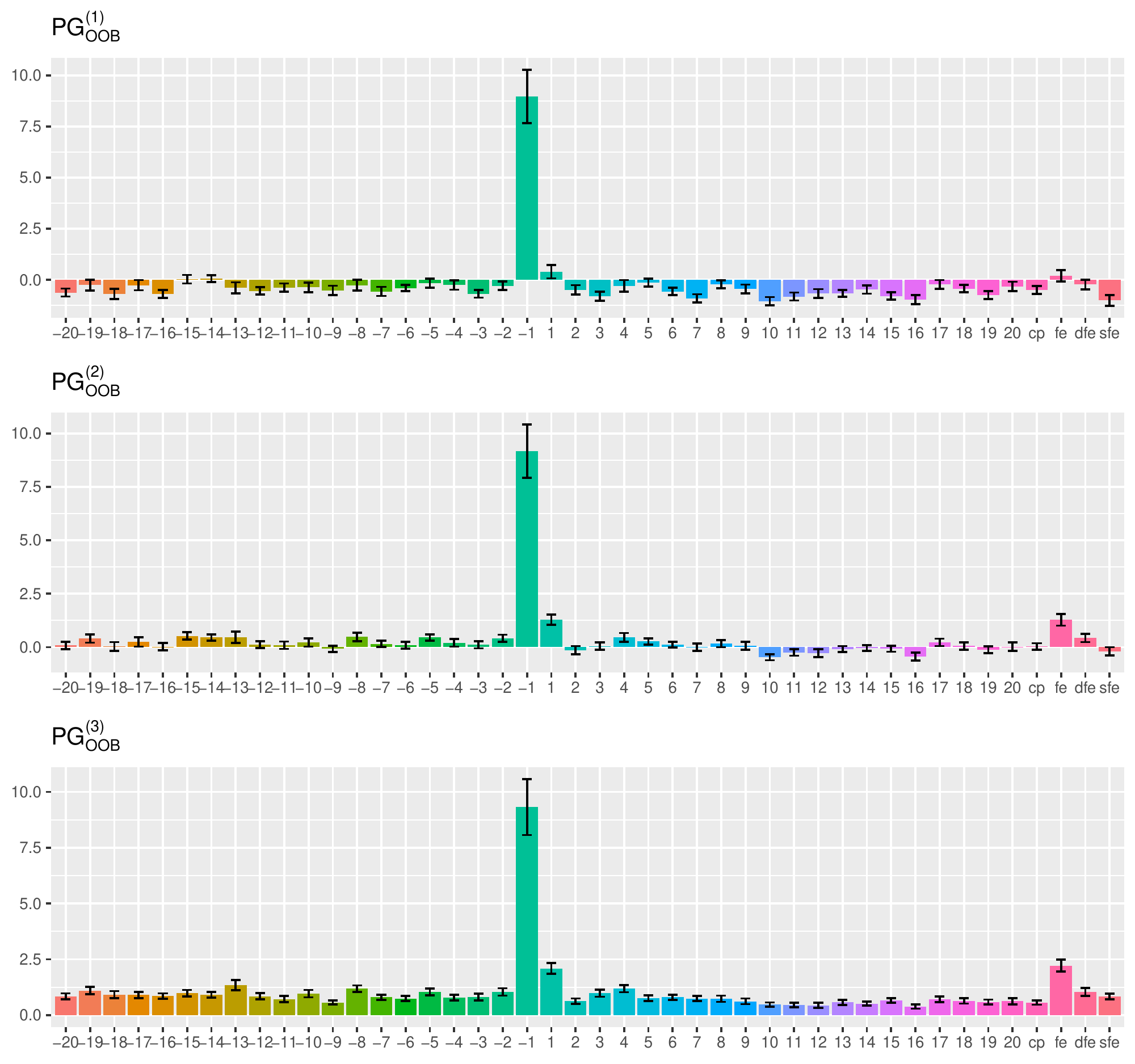}%
  \caption{Comparison of the three penalized mean decrease impurities  (PMDI) eqns. $(1)-(3)$ for for the C-to-U conversion data. Most predictors show feature importance close to $0$ but it is noteworthy to see the dependence of the scores for the continuous predictors \textit{fe,dfe} on the strength of the penalty term. The random column \textit{sfe} shows a non-zero contribution only for $PG_{OOB}^{(3)}$ . Label details as in Fig.~\ref{fig:arabidopsis1}. }
  \label{fig:arabidopsis2}
\end{figure}

As a second example, we compare importance scores for the \textit{Arabidopsis thaliana} data  \citep{cummings2004simple} which was also analyzed in \citep{Strobl2007a}.
The sample ($n=876$) contains the binary response (\textit{edit}) and the $40$ nucleotides at positions $-20$ to $20$, the codon position (\textit{cp}), the estimated folding energy (\textit{fe}) and
the difference in estimated folding energy between preedited and edited sequences (\textit{dfe}).
Only the latter $2$ predictors are continuous, the other $41$ variables are categorical with $4$ levels each.
We add an uninformative predictor (\textit{sfe}) by randomly shuffling the column \textit{fe}.
Figure~\ref{fig:arabidopsis1} shows the standard importances for a random forest model fitted to the Arabidopsis data.
Figure~\ref{fig:arabidopsis2} compares the three PDMI measures and finds only positions $-1$ and $1$ as strong predictors. while the importance of $fe, dfe$ seem to vary moderately with the exact penalty chosen.
The shuffled column \textit{sfe} is assigned a notable importance score for the default impurity decrease (MDI) on the training data  which all but disappears for the alternative measures. 

\subsection{Simulated Data \label{sec:SimulatedData}}

We replicate the simulation design used by \citep{Strobl2007a} where a binary response variable Y is  predicted from a set of $5$ predictor variables that vary in their scale of measurement and
number of categories. The first predictor variable $X_1$ is con-
tinuous, while the other predictor variables $X_2 ,\ldots, X_5$ are
multinomial with $2, 4, 10, 20$ categories, respectively. 
The sample size for all simulation studies was set to n = 120.
In the first \textit{null case}  all predictor variables and the response are sampled
independently. We woud hope that a reasonable variable importance measure would not prefer any one predictor variable over any other.
In the second simulation study, the so-called \textit{power case},
 the distribution of the response is a binomial process with probabilities that depend on the value
of $x_2$, namely $P(y=1|X_2==1)=0.35, P(y=1|X_2==2)=0.65$ .

\begin{figure}[htbp]
  \centering
  \includegraphics[width=\textwidth]{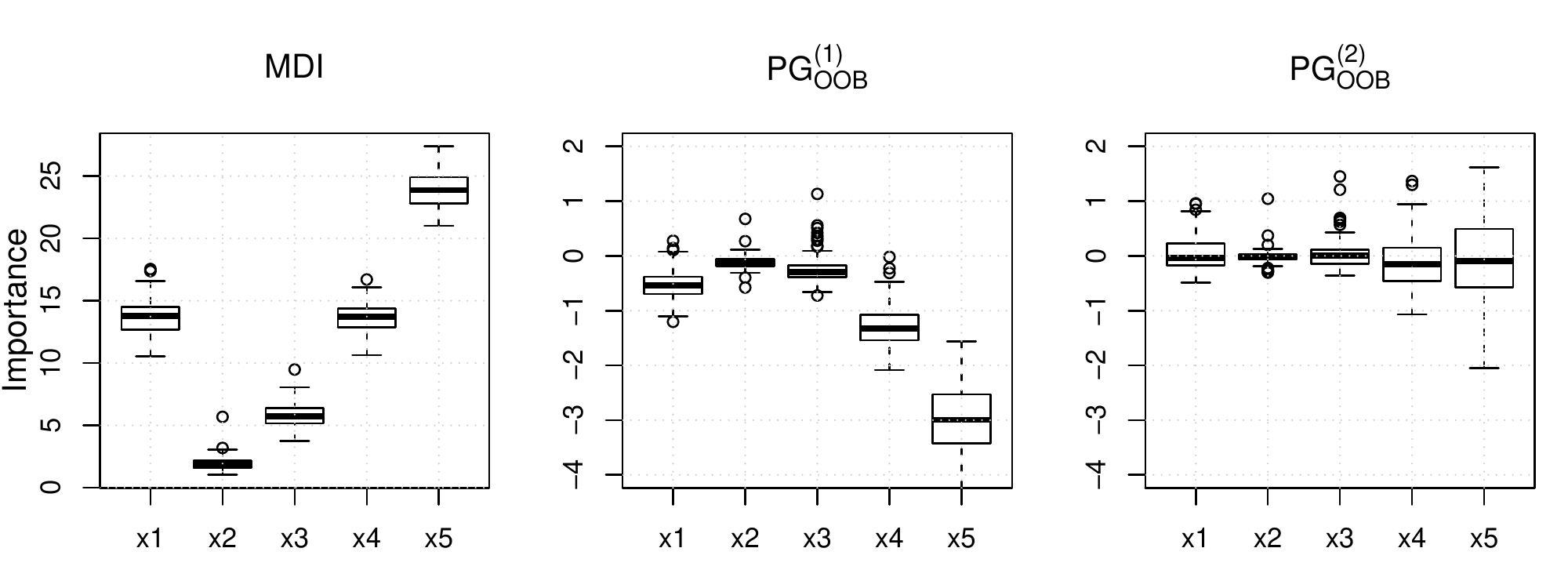}
  \caption{Results of the null case, where none of the predictor variables is informative. }
  \label{fig:NullSim}
\end{figure}

\begin{figure}[htbp]
  \centering
  \includegraphics[width=\textwidth]{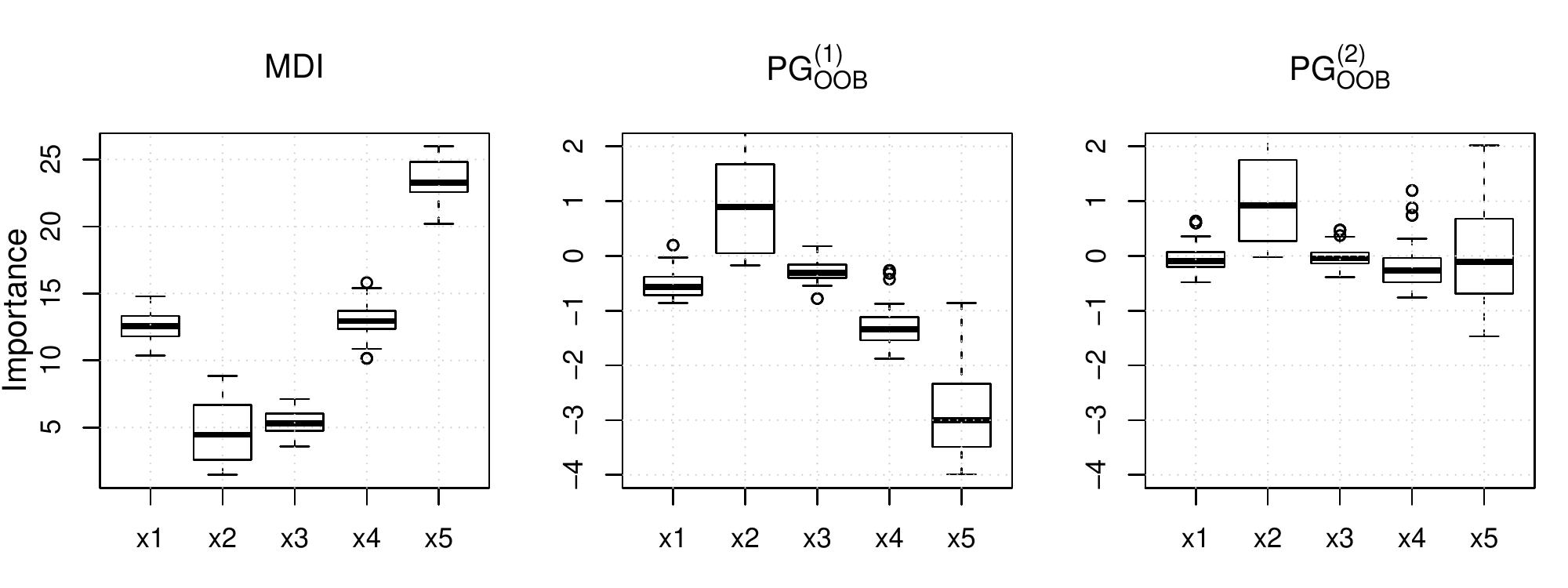}
  \caption{Results of the null case, where only $X_2$ is informative. Simulation details as in Fig.~\ref{fig:NullSim}. }
  \label{fig:PowerSim}
\end{figure}

As is evident in the left panel of Figure~\ref{fig:NullSim} the Gini importance shows a
strong preference for variables with many categories and the continuous variable confirming its well-known bias. For the sake of brevity we omit the boxplot for the permutation importance which was already discussed in \citep{Strobl2007a} and only compare eqns. (1) and (2).
Encouragingly, both methods yield low scores for all predictors. A recurring pattern of method (1) are the relatively large negative scores that depend on the cardinality of the variables. As mentioned before, it would be straightforward to clip those to zero if one was to derive no value from learning about the degree of overfitting.
More worrisome are the notable differences in the variance of the distributions for predictor variables with different scale of measurement or number of categories.
The results from the power study are summarized in Figure~\ref{fig:PowerSim}.  
MDI again shows a strong bias towards variables with many
categories and the continuous variable. At the chosen sigal-to-noise ratio it fails to identify the relevant predictor variable. In fact, the mean value for the relevant variable $X_2$ is lowest and only slightly higher than in the null case.
Both methods (1) and (2) clearly succeed in identifying $X_2$ as the most relevant feature.
While the large fluctuations of the importance scores for $X_4$ and especially $X_5$ are bound to yield moderate ``false positive'' rates and incorrect rankings in single trials. 
The negative scores of $PG_{OOB}^{(1)}$ lead to a larger separation of the signal from noise and hence more reliable correct identification of the most important variables.  

\subsection{Sample Variance Bias Correction}

Recall that the Gini index $p (1-p) = \frac{1}{N} \sum_{i=1}^{N}{(y_i - \hat{y})^2}$ is often viewed as the variance of a Bernoulli process $y_i$. 

\begin{figure}[!htb]
\minipage{0.95\textwidth}
  \includegraphics[width=\linewidth]{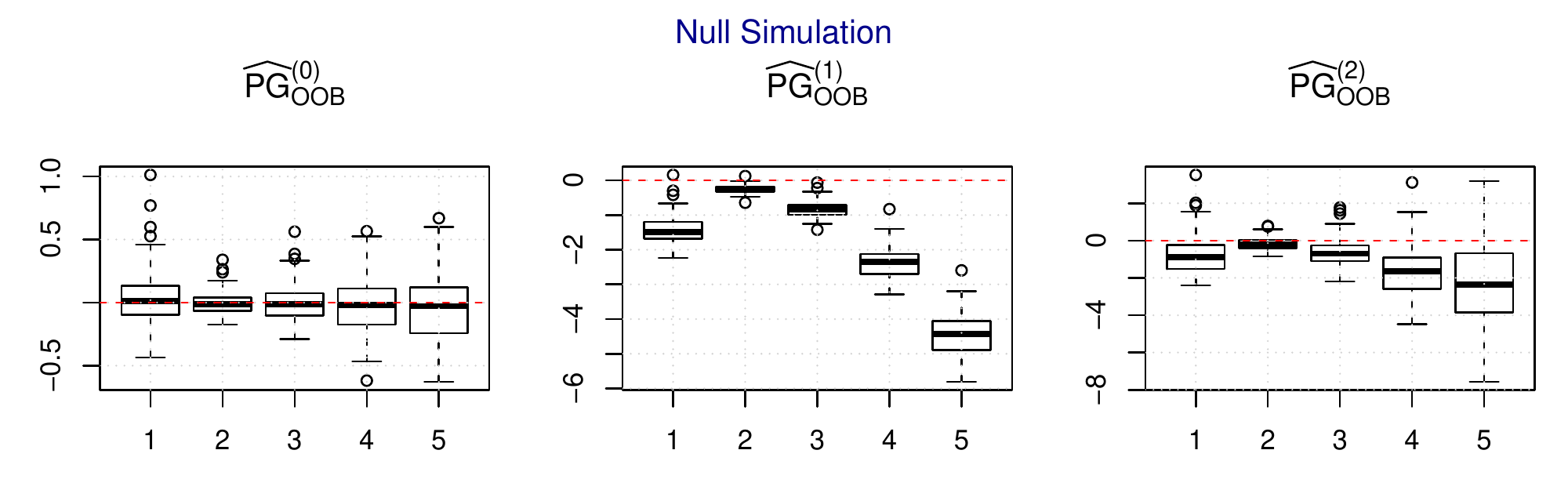}
  \endminipage\vfill
  \minipage{0.95\textwidth}
  \includegraphics[width=\linewidth]{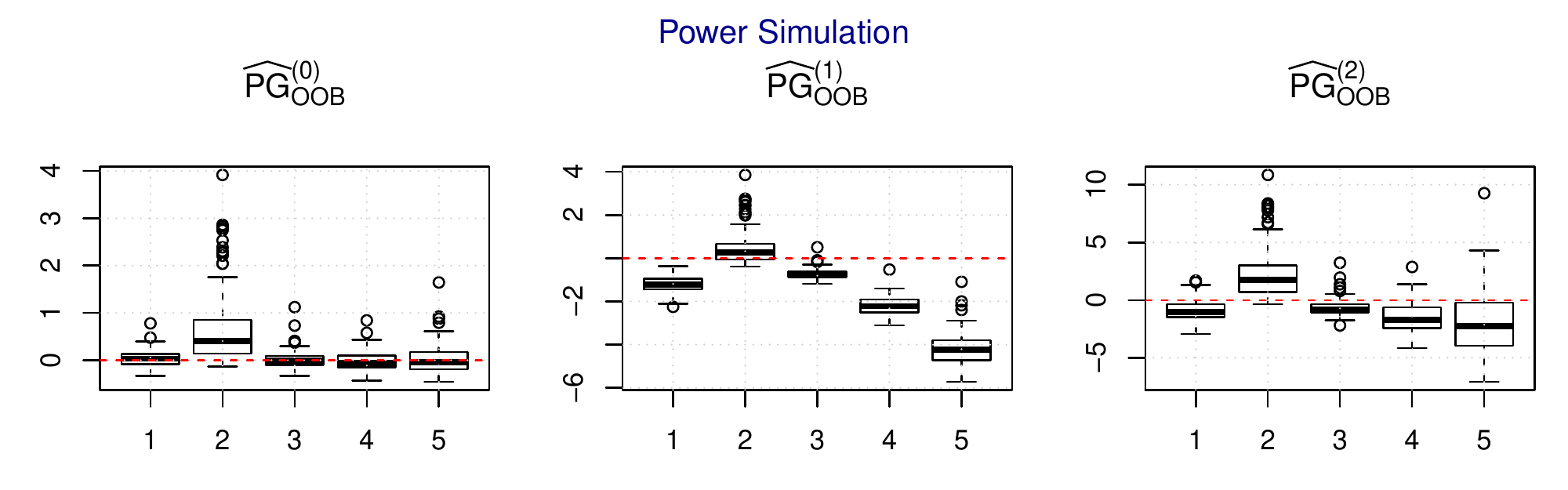}
\endminipage\vfill
\caption{Simulation results for the modified $\widehat{PG}_{oob}^{(0:2)}$. (top row) The null case, where none of the predictor variables is informative. (bottom row)  The power study, where only $X_2$ is informative.}
  \label{fig:OOBhat_PowerNull}
\end{figure}
At the same time an unbiased estimator of the variance of any data set $y_i$ is $\hat{\sigma}_y = \frac{1}{N-1} \sum_{i=1}^{N}{(y_i - \hat{y})^2}$.  We therefore propose a simple modification of Eq. (\ref{eq:PG0}):
\begin{equation}
\widehat{PG}_{oob}^{(0)}  = 2 \frac{N}{N-1} \cdot  \hat{p}_{oob} \cdot (1- \hat{p}_{oob} )  \label{eq:PG0mod}
\end{equation}
which is simply the sample version of the Bernoulli variance.
Figure \ref{fig:OOBhat_PowerNull} shows the results for the same simulations outlined in section \ref{sec:SimulatedData}. (We omitted $\widehat{PG}_{oob}^{(3)}$ because of its consistent underperformance.)
Several stark differences to Figures \ref{fig:NullSim} and \ref{fig:PowerSim} are immediately obvious: (i) $\widehat{PG}_{oob}^{(0)}$ appears unbiased (proven in the Appendix) and well-behaved, whereas (ii) $\widehat{PG}_{oob}^{(2)}$ lost its unbiased attribute and shows distorted rankings that depend on the number of categories.

\subsection{Regression}

While we leave the implementation of the regression case for a future version of the package, the idea of replacing the inbag loss (in this case MSE) with the OOB loss remains the same. 
The equivalents of Eqns. (1)-(3) will be even more straightforward as the penalty term becomes unnecessary if we replace $\bar{x}_{OOB}$ with $\bar{x}_{inbag}$: $MSE_{OOB} = \sum_{j \in OOB}{(x_j - \bar{x}_{inbag})^2}$.

\section{Discussion}

It seems somewhat surprising that the careful distinction between validation and training test has not been drawn as consistently for variable importance measures as for most other metrics in statistical modeling.
From a machine learning perspective, one would want to base feature importance scores only on the performance of an appropriate loss function on a validation set. 
Misclassification rate as a function of $\hat{p}_{OOB}, \hat{p}_{inbag})$ suffers from a discontinuity at $\hat{p}_{inbag}=0.5$. Instead we advocate the modified Gini impurity measures defined in Eqns. (1)-(3), which encode some important differences in philosophy.
For perfect agreement, $\hat{p}_{OOB} = \hat{p}_{inbag}$, all three scores reduce to the conventional Gini impurity with values between $[0, 0.5]$.
$PG_{OOB}^{(1)}$ and $PG_{OOB}^{(2)}$ will reach their maximum of $1$ when the disagreement is the highest,    $|\hat{p}_{OOB} - \hat{p}_{inbag}|=1$. 
$PG_{OOB}^{(3)}$'s philosophy is such that no constellation can yield a loss higher than $0.5$ which would reflect complete uncertainty. Hence, e.g. $\hat{p}_{OOB}=0, \hat{p}_{inbag}=1$  is no more impure than $\hat{p}_{OOB}= \hat{p}_{inbag}=0.5$.\\
The crucial difference between (1) and (2) is symmetry: $PG_{OOB}^{(2)}$ advocates an average impurity between train and test whereas $PG_{OOB}^{(1)}$ weighs inbag much less directly. 
In particular, $PG_{OOB}^{(2)} \equiv 0.5$ if either  $\hat{p}_{OOB}=0.5$ or $\hat{p}_{inbag}=0.5$ and in that case is independent of the respective other proportion.
 We have further shown that $PG_{oob}^{(2)}$ and $\widehat{PG}_{oob}^{(0)}$ are unbiased for uninformative features. In combination with the poor performance of $PG_{oob}^{(0)}$ and $PG_{oob}^{(3)}$ on simulated data, our recommendation would be to use $\widehat{PG}_{oob}^{(0)}$  or $\widehat{PG}_{oob}^{(1)}$.

Summarizing, we view the well established bias in Gini importance measures as another manifestation of overfitting in machine learning models and propose a straightforward solution to this problem. We have demonstrated its effectiveness on 2 real and one simulated data set. While our solution is clearly amenable to any tree-based method, the bias is most pronounced for deep trees, hence the focus on Random Forests.






\section{Appendix}

\subsection{Expected Values}
\noindent
The  decrease in impurity ($\Delta G$) for a parent node $m$ is the weighted difference between the Gini importance\footnote{For easier notation we have (i) left the multiplier $2$ and (ii) omitted an index for the class membership}  $G(m) = \hat{p}_m  (1- \hat{p}_m )$ and those of its left and right children:
\[
\Delta G(m) =  G(m) - \left[ N_{m_l} G(m_l) - N_{m_r} G(m_r) \right] / N_m 
\]
 
\noindent
We assume that the node $m$ splits on an \textbf{uninformative} variable $X_j$, i.e.  $X_j$ and $Y$ are independent.\\
We will use the short notation $\sigma^2_{m, .} \equiv p_{m,.} (1-p_{m,.})$ for $.$ either equal to $oob$ or $in$ and rely on the following facts and notation:
\begin{enumerate}
  \item $E[\hat{p}_{m, oob}] = p_{m,oob}$ is the ``population'' proportion of the class label in the OOB test data (of node $m$).
  \item $E[\hat{p}_{m, in}] = p_{m,in}$ is the ``population'' proportion of the class label in the inbag test data (of node $m$).
  \item $E[\hat{p}_{m, oob}] = E[\hat{p}_{m_l, oob}] = E[\hat{p}_{m_r, oob}] =p_{m,oob}$
  \item  $E[\hat{p}_{m, oob}^2]  = var(\hat{p}_{m, oob}) + E[\hat{p}_{m, oob}]^2 = \sigma^2_{m, oob}/N_m + p_{m,oob}^2$ \\
$\Rightarrow E[G_{oob}(m)]  = E[\hat{p}_{m, oob}] - E[\hat{p}_{m, oob}^2] =  \sigma^2_{m, oob} \cdot \left(1- \frac{1}{N_m}\right)$ \\
$\Rightarrow E[\widehat{G}_{oob}(m)]  =  \sigma^2_{m, oob}$
 \item $E[\hat{p}_{m, oob} \cdot \hat{p}_{m, in}] = E[\hat{p}_{m, oob}] \cdot E[\hat{p}_{m, in}] = p_{m,oob} \cdot p_{m,in}$
\end{enumerate}
Equalities 3 and 5 hold because of the independence of the inbag and out-of-bag data as well as the independence of $X_j$ and $Y$.\\

\subsubsection{$\mathbf{E(\Delta  PG_{oob}^{(0)}) \neq 0}$}
We use the shorter notation $G_{oob} = PG_{oob}^{(0)}$:
\begin{align*}
E[\Delta G_{oob}(m)] &= E[G_{oob}(m)] - \frac{N_{m_l}}{N_{m}} E[G_{oob}(m_l)] - \frac{N_{m_r}}{N_{m}} E[G_{oob}(m_r)]   \\
& = \sigma^2_{m,oob} \cdot  \left[  1- \frac{1}{N_m} - \frac{N_{m_l}}{N_{m}} \left(1- \frac{1}{N_{m_l}}\right) - \frac{N_{m_r}}{N_{m}} \left(1- \frac{1}{N_{m_r}}\right) \right]  \\
& = \sigma^2_{m,oob} \cdot  \left[  1- \frac{1}{N_m} - \frac{N_{m_l} + N_{m_r}}{N_{m}} + \frac{2}{N_m} \right]  =  \frac{\sigma^2_{m,oob}}{N_m}
\end{align*}
 We see that there is a bias if we used only OOB data, which becomes more pronounced for nodes with smaller sample sizes. This is relevant because visualizations of random forests show that the splitting on uninformative variables happens most frequently for ``deeper'' nodes.\\

\subsubsection{$\mathbf{E(\Delta  \widehat{PG}_{oob}^{(0)}) = 0}$}
The above bias is due to the well known bias in variance estimation, which can be eliminated with the bias correction (\ref{eq:PG0mod}), as outlined in the main text.
We now show that the bias for this modified Gini impurity is zero for OOB data.
As before, $\widehat{G}_{oob} = \widehat{PG}_{oob}^{(0)}$:
\begin{align*}
E[\Delta \widehat{PG}_{oob}(m)] &= E[\widehat{G}_{oob}(m)] - \frac{N_{m_l}}{N_{m}} E[\widehat{G}_{oob}(m_l)] - \frac{N_{m_r}}{N_{m}} E[\widehat{G}_{oob}(m_r)]  \\
& = \sigma^2_{m,oob} \cdot  \left[  1 - \frac{N_{m_l} + N_{m_r}}{N_{m}}  \right]  =  0
\end{align*}

\subsubsection{$\mathbf{E(\Delta  PG_{oob}^{(2)}) = 0}$}
\noindent
We can rewrite $PG_{oob}^{(2)}$ as follows:
\[
PG_{oob}^{(2)} =  \hat{p}_{oob}   +   \hat{p}_{in} - 2  \hat{p}_{oob} \cdot  \hat{p}_{in} \Rightarrow
\]
\begin{align*}
E[\Delta  PG_{oob}^{(2)}] &= E[PG_{oob}^{(2)}(m)]  - \frac{N_{m_l}}{N_{m}} E[PG_{oob}^{(2)}(m_l)] - \frac{N_{m_r}}{N_{m}} E[PG_{oob}^{(2)}(m_r)]   \\
& = p_{m,oob} + p_{m,in} - 2 p_{m,oob} \cdot p_{m,in} \\
&  - \frac{N_{m_l}}{N_{m}} \left(  p_{m,oob} + p_{m_l,in} - 2 p_{m,oob} \cdot p_{m_l,in} \right) - \frac{N_{m_r}}{N_{m}} \left(  p_{m,oob} + p_{m_r,in} - 2 p_{m,oob} \cdot p_{m_r,in} \right) \\
& = p_{m,oob} \underbrace{\left( 1- \frac{N_{m_l}}{N_{m}}  - \frac{N_{m_r}}{N_{m}} \right)}_{=0}  + (1- 2 p_{m,oob}) \cdot  \underbrace{\left( p_{m,in} - \frac{N_{m_l}}{N_{m}} p_{m_l,in}  - \frac{N_{m_r}}{N_{m}} p_{m_r,in} \right)}_{=0} \\
& = 0
\end{align*}


\end{document}